\documentclass{article}

\usepackage[preprint]{neurips_2024}

\usepackage{tabularx} 
\usepackage{geometry}
\usepackage{array}
\usepackage{wrapfig}
\usepackage{multirow}
\usepackage{booktabs} 
\usepackage[noend]{algpseudocode}
\usepackage{algorithm}
\usepackage{amsmath}
\usepackage{amssymb}
\usepackage{subcaption} 
\floatname{algorithm}{Algorithm}
\algrenewcommand\alglinenumber[1]{ #1:}
\algrenewcommand\algorithmicrequire{\textbf{Input:}}
\algrenewcommand\algorithmicensure{\textbf{Output:}}
\usepackage{algorithm}

\usepackage[pdftex]{graphicx}
\usepackage[utf8]{inputenc} 
\usepackage[T1]{fontenc}    
\usepackage{url}            
\usepackage{amsfonts}       
\usepackage{nicefrac}       
\usepackage{microtype}      
\usepackage{xcolor}         

\title{Empowering Large Language Models to Set up a Knowledge Retrieval Indexer via Self-Learning}

\author{
    \begin{tabular}{c}
    Xiang Liang$^{1,}$\hspace{-1mm} \thanks{The authors contribute equally.} \hspace{1.5mm}
    Simin Niu$^{1,}$\hspace{-0.2mm}$^{*}$ \hspace{0.5mm}
    Zhiyu Li$^{2,}$ \hspace{-1mm}\thanks{Corresponding author:lizy@iaar.ac.cn.}  \hspace{1.5mm}
    Sensen Zhang$^{1}$ \hspace{.5mm}
    Shichao Song$^{1}$ \hspace{.5mm}
    Hanyu Wang$^{1}$ \\
    Jiawei Yang$^{1}$ \hspace{.5mm}
    Feiyu Xiong$^{2}$ \hspace{.5mm}
    Bo Tang$^{2}$ \hspace{.5mm}
    Chenyang Xi$^{2}$ \\
    \end{tabular}
    \\ \vspace{.5mm}
    \begin{tabular}{c}
    $^1$Department of Computer Science, Renmin University of China \\
    $^2$Institute for Advanced Algorithms Research, Shanghai, China \\
    \end{tabular}
}

\begin{document}

\maketitle

\begin{abstract}
Retrieval-Augmented Generation (RAG) offers a cost-effective approach to injecting real-time knowledge into large language models (LLMs). Nevertheless, constructing and validating high-quality knowledge repositories require considerable effort. We propose a pre-retrieval framework named Pseudo-Graph Retrieval-Augmented Generation (PG-RAG), which conceptualizes LLMs as students by providing them with abundant raw reading materials and encouraging them to engage in autonomous reading to record factual information in their own words. The resulting concise, well-organized mental indices are interconnected through common topics or complementary facts to form a pseudo-graph database. During the retrieval phase, PG-RAG mimics the human behavior in flipping through notes, identifying fact paths and subsequently exploring the related contexts. Adhering to the principle of \textit{the path taken by many is the best}, it integrates highly corroborated fact paths to provide a structured and refined sub-graph assisting LLMs.
We validated PG-RAG on three specialized question-answering datasets. 
In single-document tasks, PG-RAG significantly outperformed the current best baseline, KGP-LLaMA, across all key evaluation metrics, with an average overall performance improvement of 11.6\%. Specifically, its BLEU score increased by approximately 14.3\%, and the \(F1_{QE}\) metric improved by 23.7\%.
In multi-document scenarios, the average metrics of PG-RAG were at least 2.35\% higher than the best baseline. Notably, the BLEU score and \(F1_{QE}\) metric showed stable improvements of around 7.55\% and 12.75\%, respectively.
Our code: \url{https://github.com/IAAR-Shanghai/PGRAG}.
\end{abstract}

\section{Introduction}
Retrieval-Augmented Generation (RAG) has garnered significant attention for its ability to integrate real-time knowledge into Large Language Models (LLMs) [1-6]. 
Nevertheless, achieving this requires the elegant construction of an efficient index [7, 8]. A well-designed index can significantly enhance retrieval speed, ensuring that LLMs quickly access relevant background knowledge, thereby reducing the delay in generating responses [9]. Additionally, it can effectively filter out irrelevant information, thereby improving the relevance of the retrieved content and making the generated results more accurate and valuable [10-12]. Constructing the index involves handling large volumes of texts, and organizing and storing them presents a challenge [13].

Traditional index construction methods typically segment texts into fixed chunks, which are then embedded and constructed into a standard index [8, 14]. While this approach is straightforward to implement, it overlooks the global context, potentially leading to incomplete information retrieval.
To address this, researchers have proposed various index structures that can be categorized into three main types:
(1) \textbf{chains} [7, 15]: retain context information to ensure continuity and completeness of information. Common implementations include overlapping chunks and progressively expanding chunks. Overlapping chunks maintain sequential coherence of information, while progressively expanding chunks provide more comprehensive context by initially retrieving small chunks and then expanding to larger information blocks;   
(2) \textbf{trees} [16, 17]: organize hierarchical relationships between text blocks by summarizing similar blocks layer by layer, capturing structured information within the document;
(3) \textbf{graphs} [18]: represents complex relationships between blocks by establishing similarity-based connections between text blocks.

\begin{figure}
  \centering
  \includegraphics[width=\linewidth]{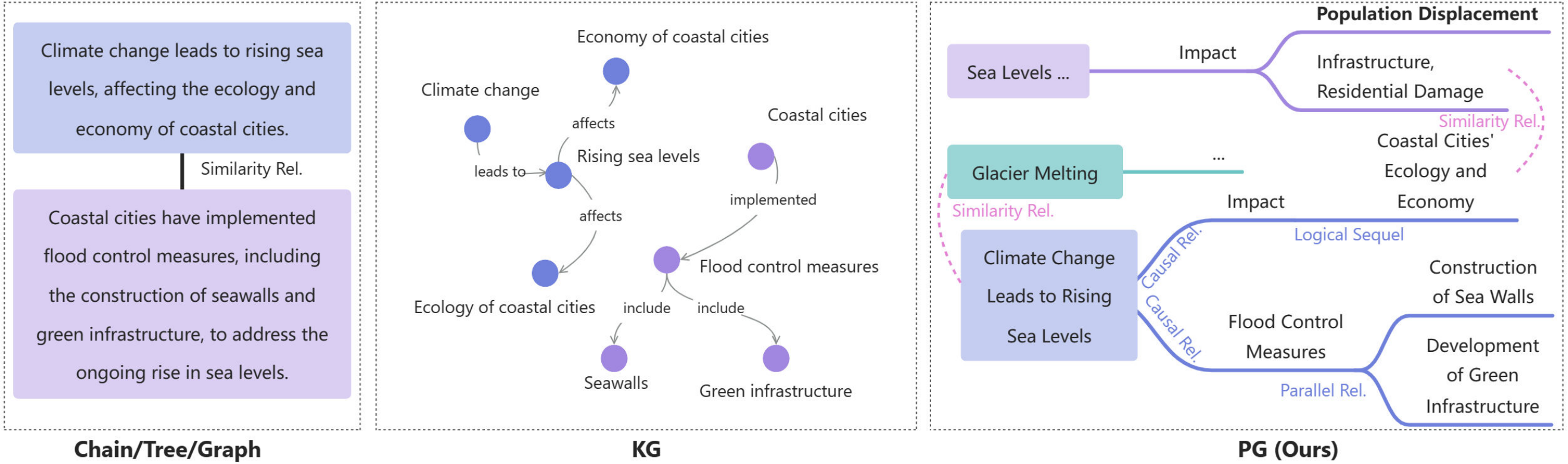}
  \caption{\textbf{Motivation}. Existing RAG suffers from several issues: (1) \textbf{limited retrieval granularity}: Chain/Tree/Graph-RAG typically use a fixed granularity, such as sentences. KG-RAG employs type-preset structures, such as entities and triples; (2) \textbf{restricted relationships}: similarity is often utilized to generalize various logical relations, and the relations in KGs necessitate strict definitions.}
  \label{fig:motivation}
\end{figure}

Despite simplicity and generality, the indexing methods above have limitations in knowledge representation capabilities. For example, KGP [18] attempts to represent relationships between blocks, such as parallel relations and sequential reasoning relations, using simple similarity measures to support complex queries.
Additionally, most of these methods are based on the blocks themselves, making them susceptible to the influence of blocks containing irrelevant content. To mitigate this, [19] indexes at the fine-grained proposition level, ensuring that each proposition contains all necessary context information from the text, thereby reducing the interference of irrelevant information and improving the quality of knowledge. However, this method lacks the construction of relations between knowledge points, thus making complex reasoning difficult.
Methods such as T-RAG, KG-RAG [10-12, 20] attempt to construct indexes on high-quality entity trees or knowledge graphs (KGs), providing context with high knowledge density and enhanced reasoning capabilities. However, the extraction of entities and the construction of relations are inherently challenging, and these methods often require preset schemas, thereby limiting their generality. For example, in Figure \ref{fig:motivation}, if the preset relation in the KG is "\textit{effects}", it may not handle semantically consistent but differently expressed relations like "\textit{affects}", even though they are similar in meaning.
We have summarized these RAG methods in Table \ref{tab:rw}. For more detailed information, refer to Appendix \ref{Related Work}.

To balance knowledge representation capabilities and generality, we adopted a self-learning approach for LLMs to develop an efficient knowledge retrieval indexer.
Specifically, we treat LLMs as students, providing them with raw reading materials and encouraging it to self-learn and create mind maps. By reading these materials, LLMs organizes the knowledge within the documents, establishing a hierarchical index which not only avoids noise but also conveys meaning more clearly. Subsequently, we establish connections between similar or complementary pieces of knowledge, allowing the index to span multiple documents. 
During the knowledge retrieval phase, we first generate Key Points (KPs) which are similar to [21] as pseudo-answers to assist the retriever in exploring knowledge paths on the pseudo-graph. Simultaneously, we optimized the step-by-step evaluation of nodes and path prediction during the depth-first search (DFS) process. The designed pseudo-graph retrieval (PGR) algorithm transfers the processes of exploration (walking) and evaluation (discrimination) to efficient matrices, thereby accelerating the localization of evidence. Finally, we assemble the most significant knowledge paths into a structured context, which is then fed into LLMs along with the query to generate the response.
We highlight our contributions as follows:

\begin{itemize}

    \item \textbf{Balancing Generalization and Knowledge-based Indexing Paradigm}: Leveraging the general learning capabilities of large language models (LLMs) to autonomously organize document content, thereby creating hierarchical indexes, which reduces preset limitations, internalizes external knowledge, and achieves knowledge alignment.

    \item \textbf{Cross-document Knowledge Construction}: Establishing cross-document knowledge relationships by connecting concise and well-organized mental indexs based on common or complementary contents, which enhances the organization and accessibility of knowledge indexes, thereby improving the model's ability to handle complex queries.

    \item \textbf{Structured Supporting Evidence}: Using LLMs to transform raw queries into KPs, thereby stimulating the utilization of the model's internal implicit knowledge, which involves retrieving and integrating highly relevant knowledge paths related to KPs, building a hierarchical context that can be directly processed by the model without additional interpretation.
\end{itemize}

\section{Method}
\begin{figure}
  \centering
  \includegraphics[width=\linewidth]{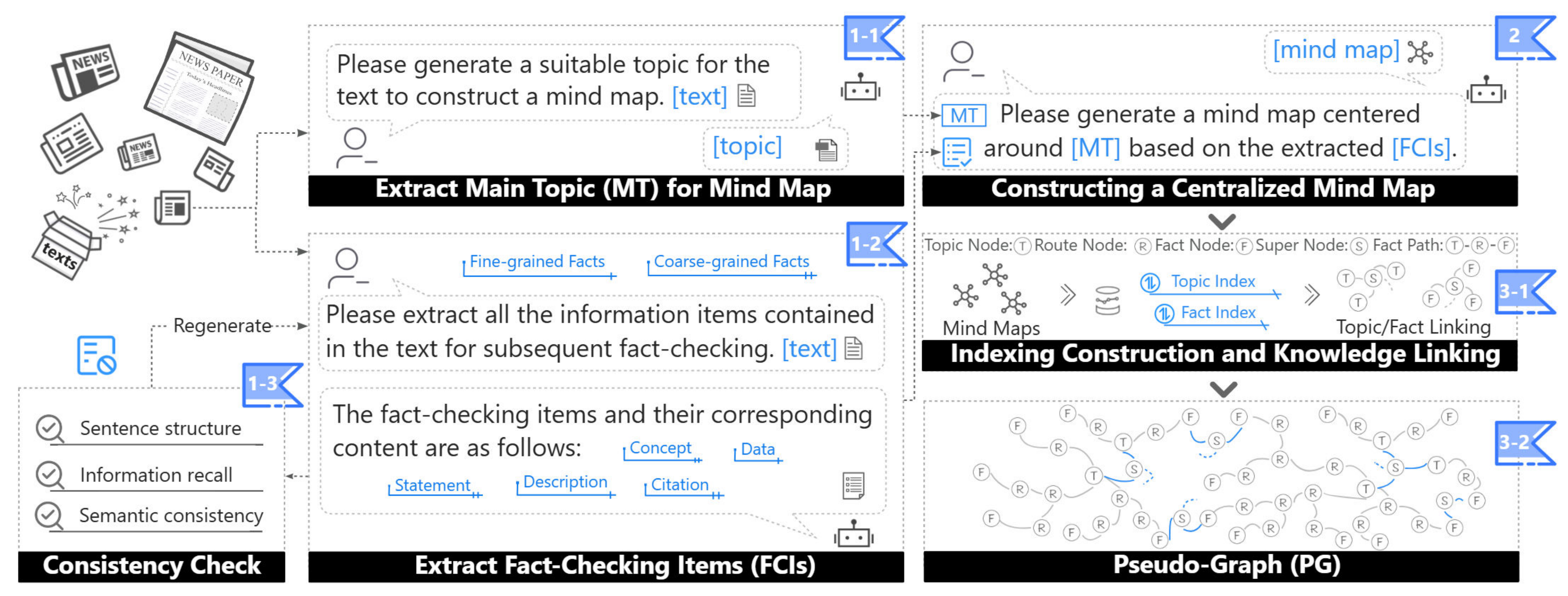}
  \caption{\textbf{Pseudo-Graph Construction Overview}. (1) We first extract \textit{FCIs} to convert a text into loosely structured evidence entries and perform \textit{consistency checks} on \textit{FCIs}.
(2) We input the verified \textit{FCIs} along with their \textit{MT} into LLMs to generate a \textit{mind map}.
(3) We perform knowledge fusion at the topic and fact levels, constructing \textit{super nodes} and cross-document links to form a \textit{pseudo-graph}.}
\label{fig:pg}
\end{figure}
\subsection{Preliminaries}
\textbf{Pseudo-Graph.} Similar to traditional graph definitions [22], let \( \mathcal{PG} = (\mathcal{M}, V_s, E_s) \) represent the \textit{pseudo-graph} composed of mind maps \(\mathcal{M}\) extracted from a set of documents \( D \), along with super nodes \(V_s\) and super relations \(E_s\). A mind map \(M\in\mathcal{M}\) connects \textit{topic}, \textit{routes}, and \textit{facts} through \textit{navigational lines}, forming rich inter-document relationships:
(1) \textbf{topic entities} \( V_t \) are the set of all mind map topics, and the topic provide a clear starting point for organizing information in the mind map.
(2) \textbf{fact entities} \( V_f \) offer detailed data or descriptions directly related to the topic;
(3) \textbf{route entities} \( V_r \) serve as intermediate information nodes between topic and fact entities, helping to elucidate and guide the knowledge transfer paths;
(4) \textbf{navigational lines} \( E_r \) express various logical relationships between entities, guiding the flow of knowledge.
The set of node features \( X \) includes the characteristics of all entities, with each feature \( x_i \in X \) corresponding to the knowledge content of node \( v_i\in V_{t,r,f} \).
For any given topic entity \( v_t \in V_t \), there is a corresponding mind map \( M_{t} \in \mathcal{M} \), defined as follows:
\begin{equation}
    M_{t} = \{ C_{t}^i \} = \left\{ (v_t, e_r^1, v_r^1, e_r^2, v_r^2, \ldots, e_r^n, v_r^n, e_r^{n+1}, v_f) \mid e_r \in E_r, v_r \in V_r, v_f \in V_f \right\},
\end{equation}
where \( C_{t}^i \) represents fact paths, i.e., routing chains. Each fact path starts from the topic entity \( v_t \), passes through a series of route entities \( v_r^i \) connected by navigational lines \( e_r^i \), and eventually leads to the corresponding fact entity \( v_f \).
Moreover, the super nodes \(V_s\) consist of super-topic nodes \(V_{st}\) and super-fact nodes \(V_{sf}\), connected by super relations \(E_s\) between topic nodes \(v_t\) or fact nodes \(v_f\). For example, \((v_t^1, e_s, v_t^2), e_s\in E_s\) indicates that topic \(v_t^1\) and topic \(v_t^2\) share a thematic similarity, facilitating cross-document information integration and knowledge discovery.

\textbf{Knowledge Embedding Learning.} Traditional RAG methods primarily adopt independent embedding approaches [8, 14], lacking the modeling of relationships between pieces of knowledge. We propose a node embedding learning method based on fact paths, where the embedding vector \( \text{emb}(v_i) \) of each node \( v_i \) represents the concatenated fact chain formed by the attributes of all nodes in the path from the starting topic \( v_t \) to the current node \( v_i \):
\begin{equation}
    \text{emb}(v_i) = \text{emb}(\text{concat}(x_1, x_2, \ldots, x_i))\mid v_i \in \{V_t, V_r, V_f\},
\end{equation}
where \( x_i \) is the attribute of \( v_i \), i.e., the knowledge text, and \(\text{concat}(\cdot)\) represents the concatenation operation of the texts. In this paper, we use spaces to concatenate different texts. The \(\text{emb}(\cdot)\) denotes the embedding process, and we use pre-trained models \textit{sentence transformers} [23] to capture the structured path information formed by all informational nodes from the topic to the current node.
\subsection{Problem Statement}
Given a query \( \vec{q} \), the generator models a conditional distribution \( p(\vec{o}|\vec{q}) \) to generate an output response \( \vec{o} \). To generate knowledge-based responses, RAG first retrieves relevant knowledge contexts \( \mathcal{Z}\), and then generates the response \(\vec{o}\) based on contexts \( \mathcal{Z}\). Ideally, the generator and the retriever should be jointly optimized to find an optimal context \( \mathcal{Z} \), i.e., maximizes the probability \(
\max_{\mathcal{Z}} p(\mathcal{Z}|\vec{q}) \cdot p(\vec{o}|\vec{q}, \mathcal{Z})
\), where \( p(\mathcal{Z}|\vec{q}) \) and \( p(\vec{o}|\vec{q}, \mathcal{Z}) \) represent the probabilities of the retriever and the generator, respectively. In practice, due to computational resource constraints and optimization difficulties, we typically choose to optimize the retriever separately to retrieve a sub-graph \( \mathcal{Z} \subseteq \mathcal{PG} \) composed of fact paths \( c_{fp} \in C \), which explicitly provide knowledge relevant to the current query \( \vec{q} \).

\subsection{Pseudo-Graph Construction}
As shown in Figure \ref{fig:pg}, the construction of the pseudo-graph mainly involves three core steps: (1) extraction and verification of fact-checking items (FCIs), (2) generation of mind maps, i.e., organizing knowledge within a single document, and (3) construction of inter-document knowledge relationships.

\textbf{Construction of Mind Maps.} PG-RAG first transforms the original text \(\mathbf{d}_{orig} \in D\) into fact-checking texts \( \mathbf{d}_{fci} = [d_{fci}^1, \ldots, d_{fci}^n] \) generated by LLMs. During knowledge extraction, LLMs focus on identifying fact \( d_{fci}^i \) that can verify the original text. This targeted fact-checking extraction approach concentrates on verifiable facts that can directly support or refute the original statements, thereby minimizing misleading information due to missing context or misinterpretations. FCIs may involve fine-grained numerals, dates, and locations, but also include coarse-grained macro facts, such as viewpoints or policy content. Compared to direct extraction, our pre-transformation process retains facts more comprehensively.
Next, we perform consistency checks on the extracted FCIs using the evaluation function \( f \) which combines BERT-Score \( f_{bs} \) and ROUGE-L \( f_{rl} \) and define two thresholds, \(\theta_{bs}\) and \(\theta_{rl}\), for them:
\begin{equation}
    f(\mathbf{d}_{orig}, \mathbf{d}_{fci}) = (f_{bs}(\mathbf{d}_{orig}, \mathbf{d}_{fci}) \geq \theta_{bs]}) \land (f_{rl}(\mathbf{d}_{orig}, \mathbf{d}_{fci}) \geq \theta_{rl}),
\end{equation}
where \(\land\) denotes the logical "\textit{and}" operation. Then, we get the output as follows:
\begin{equation}
    \mathbf{d}_{fci} \leftarrow \mathbf{d}_{fci} \cdot \mathbb{I}(f(\mathbf{d}_{orig}, \mathbf{d}_{fci})) + \operatorname{<regenerate>} \cdot \mathbb{I}(\neg f(\mathbf{d}_{orig}, \mathbf{d}_{fci})),
\end{equation}
where \(\mathbb{I}(\cdot)\) is the indicator function. Thus, if the evaluation result of is true, the output fact-checking text \( \mathbf{d}_{fci} \) passes the verification; otherwise, the output is the instruction "<regenerate>" and the fact extraction is performed again.
We further use LLMs to perform the second transformation on the verified FCIs to form a hierarchical mind map \(
M_i\) around the topic, i.e.,
\(
M_i = \text{LLM}(\mathbf{d}_{fci}^i, x_t^i),
\)
where \( x_t^i \in X \) is the attribute of the topic node \( v_t^i \in V_t \). 

\textbf{Establishment of Cross-Document Relationships.} 
We integrated the collected mind maps \( \mathcal{M} \) by constructing clusters and linking knowledge, thereby establishing cross-document associations to form a structured pseudo-graph network. During the clustering phase, we first performed knowledge embedding learning on the nodes and calculated the clustering score \( sim^{cluster} = \text{sim}( \text{emb}(v_i), \text{emb}(v_j) ) \). Based on the similarity \( sim^{cluster} \), the similar node sets were grouped into clusters, with each cluster assigned a super node as a connection point. Specifically, we set similarity thresholds of 0.92 and 0.98 for the topic nodes \( V_t \) and the fact nodes \( V_f \), respectively. 
Only nodes exceeding the thresholds were clustered together.
Once clusters were formed, the originally isolated mind maps \( \mathcal{M} \) was connected through super nodes \( V_s \), allowing similar topics or complementary facts to be linked, thereby enhancing the navigability and accessibility of the information within the network. The specific algorithm details and the case of pseudo-graph generation are provided in Appendices \ref{knowledge fusion algorithm} and \ref{Case Study}, respectively.

\subsection{Knowledge Recall via Pseudo-Graph Retrieval}
\begin{figure}
  \centering
  \includegraphics[width=\linewidth]{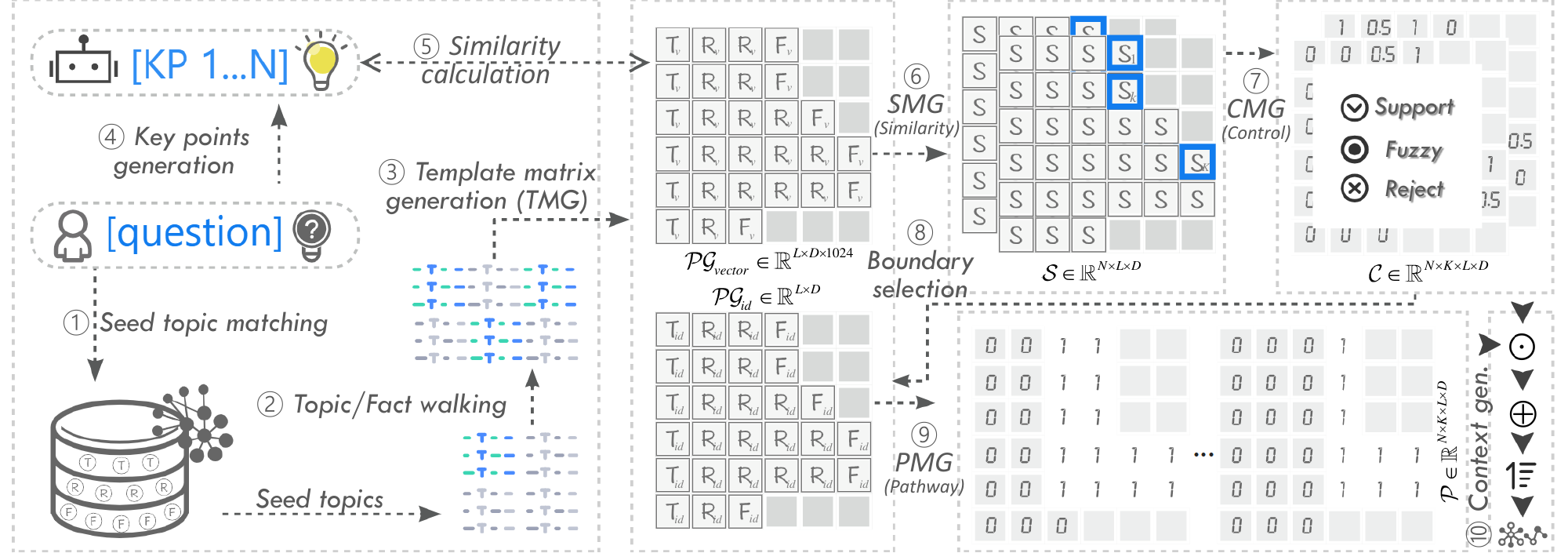}
  \caption{\textbf{Pseudo-Graph Walking Overview.} (1) Query the \textit{pseudo-graph} database to locate \textit{seed topics}.
(2) Explore nodes linked to \textit{seed topics} to form \textit{candidate nodes}.
(3) Load ids and vectors of the \textit{candidate nodes} into \textit{template matrices} (TMs).
(4) Generate \textit{KPs} to assist retrieval.
(5) Get \textit{similarity matrices} (SMs) based on \textit{KPs} and \(\mathcal{PG}_{vector}\).
(6) Select top fact paths as \textit{seed nodes}.
(7) Create \textit{control matrices} (CMs) to assess the contribution value of \textit{candidate nodes} to \textit{seed nodes}.
(8) Select \textit{boundaries} to determine the reachable area from \textit{seed nodes}.
(9) Set values within \textit{boundaries} to 1, generate \textit{pathway matrices} (PM) for \textit{seed nodes}.
(10) Multiply CMs by PMs to generate Staining Matrices (SMs), which indicate the contribution values of \textit{candidate nodes} accessible from \textit{seed nodes}. Sum all SMs to produce an Aggregate Matrix (AM), representing the overall contribution of \textit{candidate nodes} to query. Integrate the top rows (fact paths) from AM into structured context \(\mathcal{Z}\).
}
\label{fig:pgr}
\end{figure}
\textbf{Template Matrices Preload.} When humans search for information, they typically use clear "\textit{clues}" to guide their search in an orderly and purposeful manner. Once these key points are identified, they further refine the evidence by selecting the relevant context. Our PGR algorithm mirrors similar retrieval logic. As shown in Figure \ref{fig:pgr},
first, among broad topics \( V_t \), we use similarity to filter out seed topic nodes \( V_t^{seed} \subset V_t \) that are most relevant to the query. Then, by exploring topic nodes \( V_t^{con} \) associated with them, we merge and obtain candidate topic nodes \( V_t^{candi} = \{V_t^{seed} \cup V_t^{con}\} \).
Specifically, we achieve topic walking through super-topic nodes \( V_{st} \) to obtain related topic nodes, i.e., \( V_t^{con} \rightarrow V_{st} \leftarrow V_t^{seed} \). Additionally, we perform fact walking through super-fact nodes \( V_{sf} \) on fact nodes \( V_f^{seed} \stackrel{*}{\leftarrow} V_t^{seed} \) connected to seed topics \( V_t^{seed} \) to explore other possibly related topics, i.e., \( V_t^{con} \stackrel{*}{\rightarrow} V_f^{con} \rightarrow V_{sf} \leftarrow V_f^{seed} \).
Within the defined range of topics \( V_t^{candi} \in V^{candi} \), where candidate nodes \( V^{candi} \) include candidate topics \( V_t^{candi} \) itself and all nodes associated with it. Then, we use Key Points (KPs) to locate the fact nodes \( V_f^{top} \subset V^{candi} \) with the top similarity in parallel, and select them as seed nodes \( V_f^{seed} \), due to the fact node encode the entire path content from topic to fact. The generation prompt and explanations for KPs can be found in Appendix \ref{Case Study}.
Next, starting from seed nodes \( V_f^{seed} \), we perform an upward DFS to simulate the behavior of searching for context. We optimized the natural selection of gradually performing DFS exploration from the seed nodes, as each step's decision depends on the previous step's evaluation, significantly increasing the total traversal time. In detail, we preloaded the IDs required for walking and the vectors needed for evaluation into template matrices (TMs) which are used for the parallel execution of importance evaluation and selection of candidate nodes \( V^{candi} \). Specifically, the ID template matrix \(\mathcal{PG}_{id}\) records the IDs of an entire path from the root to the leaf in each row. The vector template matrix \(\mathcal{PG}_{vector}\) stores the knowledge embeddings at the corresponding positions.

\textbf{Pathway Matrix Generation.} In TMs, tracing from leave (fact) to the root (topic) can be achieved through reverse traversal, i.e., moving from the right to the left of the row. To continue exploring reachable child node paths from any parent node \( v \in {V}_{t,r} \), one needs to find positions in the ID matrix \(\mathcal{PG}_{id}\) where the IDs match the parent node ID, i.e., the co-parent node indices \({ind}_{cp}\).
By traversing right from these positions, one can find traversal paths from the parent node as follows:
\begin{equation}
    {trvP}_{cp} = \{ \mathcal{PG}_{id}[i: k] \mid k \geq {ind}_{cp}[1], \mathcal{PG}_{id}[i, k] \neq -1 \}.
\end{equation}
Concretely, in TM \(\mathcal{PG}_{id}\), all traversal paths with the same parent node appear in consecutive rows. Therefore, a DFS from any node \( v^{start} \) to a parent node \( v^{end} \) can be transformed into determining the boundaries of the consecutive rows containing the parent node \( v^{end} \).
First, the indices of the co-parent nodes are determined as
\(
{ind}_{cp} = \{(i, j) \mid \mathcal{PG}_{id}[i, j] = \text{id}(v^{end}) \}.
\)
The left boundary \(
bnd_{l} = {ind}_{cp}[1]
\) is defined by the column index of the parent node.
The top and bottom boundaries are defined by the starting and ending row coordinates of the co-parent nodes:
\begin{equation}
    (bnd_{t}, bnd_{b}) = ( \min(i), \max(i) \mid \mathcal{PG}_{id}[i, bnd_{l}] = \text{id}(v^{end}) ).
    \label{eq:6}
\end{equation}
Next, traverse right from these positions in parallel until IDs are not empty to establish right boundary:
\begin{equation}
    bnd_{r} = \min \left\{ k \mid \mathcal{PG}_{id}[i, k] \neq -1, i \in [bnd_{t} : bnd_{b}] \right\}.
    \label{eq:7}
\end{equation}

Then, set values within the region to 1 to construct a pathway matrix from a seed node as follows:
\begin{equation}
    PM_{i,j}^{v^{seed}} = (bnd_t \leq i \leq bnd_b) \land (bnd_l \leq j \leq bnd_r).
    \label{eq:8}
\end{equation}

\textbf{Control Matrix Generation.} To provide comprehensive and detailed context for high-quality seed nodes, we impose controls on candidate nodes. Specifically, for each seed node \( v^{seed} \), we calculate contribution values of candidate nodes \( V^{candi} \) to the seed node and generate a Control Matrix \( CM^{v^{seed}} \). By comparing the candidate node similarity \( {sim}^{candi} \) with the seed node similarity \( {sim}^{seed} \), we set three control conditions:
    (1) \textbf{support:} when the similarity difference meets the support threshold \( \theta_s = 0.03 \), the contribution value is \( 1 \cdot {sim}^{candi} \), indicating that the current node highly supports the seed node in answering the query;
    (2) \textbf{fuzzy:} when the difference only meets the fuzzy threshold \( \theta_f = 0.05 \), the contribution value is \( 0.5 \cdot {sim}^{candi} \), indicating that the current node somewhat supports the seed node;
    (3) \textbf{reject:} the contribution value is 0.

Next, starting from seed node \( v^{seed} \), we traverse to left, calculating and recording node contributions in \( CM^{v^{seed}} \) until reaching the left boundary  \( bnd_{l} \) required for the PM where the contribution value changed from 1 to 0. Then, starting from \( bnd_{l} \), we traverse right in parallel for other rows, calculating and recording the contributions in \( CM^{v^{seed}} \), stopping when we first encounter the contribution value 0, i.e., not expanding subsequent child nodes, ultimately generating a Control Matrix \( CM^{v^{seed}} \).

\textbf{Context Generation.} For each query \( q \), we concurrently obtain multiple sets of Control Matrices \( CM^{V^{seed}} \) and Pathway Matrices \( PM^{V^{seed}} \), starting from the seed nodes using the corresponding KPs. We perform element-wise multiplication of these matrices to obtain the contribution values of candidate nodes reachable from the seed nodes, known as Staining Matrices (SMs).
Next, we sum the SMs to form the Aggregate Matrix (AM), which records total contribution values of candidate nodes for the query \( q \), considering multiple KPs.
Finally, we select the rows (fact paths) with the top total contribution values, integrate their texts into the structured context \(\mathcal{Z}\) into LLMs for response:
\begin{equation}
    \mathcal{Z} = \text{topK}({\sum}_{v^{seed} \in V^{seed}} CM^{v^{seed}} \odot PM^{v^{seed}}).
    \label{eq:9}
\end{equation}
The complete process of PGR is shown in Algorithm \ref{alg:structured_context}.
\begin{algorithm}[H]
\caption{Knowledge Recall via Pseudo-Graph Retrieval Algorithm.}
\label{alg:structured_context}
\begin{algorithmic}[1]
\Require A query vector \( \vec{q} \), the ID template matrix \(\mathcal{PG}_{id}\), and the vector template matrix \(\mathcal{PG}_{vector}\).
\Ensure A structured context \( \mathcal{Z} \).

    \State Use LLMs to generate the corresponding key points \(\{\vec{q}_{KP}\}\).
    \For{each \(\vec{q}_{KP}\)}
        \State Compute the similarity matrix:
        \(
        SM = \vec{q}_{KP} \odot \mathcal{PG}_{vector}.
        \)
        \State Select seed nodes \( V^{seed} \) from \( SM \).
    \EndFor

    \For{each seed node \( v^{seed} \)}
        \State Traverse left in \( SM \) to calculate contribution values and find the left boundary \( bnd_{l} \).
        \State From \( bnd_{l} \), traverse right to calculate contribution values to form the control matrix \( CM^{v^{seed}} \).
        \State Extract the path matrix \( PM^{v^{seed}} \) from \(\mathcal{PG}_{id}\) using Eqs. \ref{eq:6}, \ref{eq:7}, and \ref{eq:8}.
    \EndFor
    
    \State Compute \( \mathcal{Z} \) using Eq. \ref{eq:9}.
\State \Return The structured context \( \mathcal{Z} \).

\end{algorithmic}
\end{algorithm}

\section{Experiment}
\begin{table}[t]
\centering
\caption{Performance (\%) of different baselines. The best and runner-up are in bold and underlined. }
\resizebox{\textwidth}{!}{
\begin{tabular}{@{}c|cccc|cccc|cccc@{}}
\toprule
\multirow{2}{*}{\textbf{Method}}                                  & \multicolumn{4}{c|}{\textbf{1-Document}}       & \multicolumn{4}{c|}{\textbf{2-Document}}        & \multicolumn{4}{c}{\textbf{3-Document}}        \\  
                                                                  & Bleu           & Rouge-L        & BertScore     & \(F1_{QE}\)             & Bleu           & Rouge-L        & BertScore      & \(F1_{QE}\)             & Bleu           & Rouge-L        & BertScore      & \(F1_{QE}\)             \\ \midrule 
w/o RAG                                                            & 41.37          & 59.59          & 81.52             & 55.54              & 21.82           & 35.52          & 83.92              & 37.96              & 20.08           & 32.55          & 84.27              & 34.72              \\  
BM25                                                              & 39.91          & 57.33          & 83.36         & 59.3           & 24.61          & 38.31          & 86.86          & 47.63          & 20.98          & 34.33          & 87.02          & 42.01          \\  
DPR                                                               & 39.76          & 57.24          & 83.81         & 60.41          & 22.75          & 37.25          & 87.16          & 48.87          & 21.05          & 35.04          & 87.81          & 45.18          \\ 
Hybrid                                                          & 39.67          & 57.38          & 84.06         & 60.44          & 24.03          & 38.43          & 87.3           & 51.11          & 21.35          & 35.34          & 87.66          & 45.54          \\ 
\begin{tabular}[c]{@{}c@{}}Hybrid+Rerank\end{tabular}       & 40.63          & 58.26          & 84.68         & 62.81          & 24.53          & 38.91          & 87.89          & 52.08          & 21.74          & 35.88          & {88.21}    & 46.54          \\ 
\begin{tabular}[c]{@{}c@{}}Tree Traversal\end{tabular} & 39.25          & 57.14          & 83.21         & 59.22               & 23.08          & 38.25          & 87.83          & 47.03          & 21.01          & 34.72          & 87.28          & 43.51        \\ 
\begin{tabular}[c]{@{}c@{}}Collapsed Tree\end{tabular} & {39.74}    & {57.50}     & {83.37}   & {60.02}     & 23.17          & 37.76          & 86.94          & 46.41          & 21.14          & 34.52          & 87.15          & 43.46      \\ 
\begin{tabular}[c]{@{}c@{}}Graph-RAG\end{tabular} & 40.32          & 58.39          & 84.38         & 61.59          & 23.32          & 38.62          & {87.91}    & 51.93          & 21.87          & {36.60}    & 87.44          & 46.98          \\ 
KGP-LLaMA                                                         & 41.47          & 58.72          & 85.36         & 63.81          & {25.53}    & {40.88}    & 87.39          & \underline{52.37} & {22.63}    & 36.16          & 88.02          & {47.10}     \\ \midrule \begin{tabular}[c]{@{}c@{}}\textbf{PG}\\ \end{tabular}           & {47.94} & {64.46} & {87.46} & {69.61} & \underline{27.98} & {40.85} & {86.90} & {50.87}    & \underline{25.08} &{36.86} & {86.98} & {49.57} \\
\begin{tabular}[c]{@{}c@{}}\textbf{PG w. KPs}\\ \end{tabular}           & \underline{53.07} & \underline{66.64} & \underline{88.40} & \underline{71.14} & {26.89} & \underline{41.69} & \underline{88.03} & {52.24}    & {24.88} & \underline{38.78} & \textbf{88.56} & \underline{51.38} \\

\begin{tabular}[c]{@{}c@{}}\textbf{PG-RAG}\\ \end{tabular}           & \textbf{55.73} & \textbf{69.68} & \textbf{89.78} & \textbf{79.66} & \textbf{29.19} & \textbf{43.08} & \textbf{88.09} & \textbf{55.71}    & \textbf{26.97} & \textbf{39.19} & \underline{88.31} & \textbf{52.83} \\\bottomrule
\end{tabular}
 }
\label{tab:retriever}
\end{table}
\subsection{Datasets}
In this experiment, we selected three Question Answering (QA) datasets from the CRUD-RAG benchmark [24] to assess the performance of the RAG system in knowledge-intensive applications. These datasets evaluate the model's ability to answer fact-based questions and its capacity to reason using information from multiple documents:
\textbf{1-Document QA} focuses on fact-based question answering, examining the model's ability to precisely locate and extract relevant information.
\textbf{2-Document QA} tests whether the model can utilize and integrate information from two documents for reasoning.
\textbf{3-Document QA} involves questions that require the model to simultaneously synthesize information from three documents, further increasing the complexity of the task and demanding deeper understanding and analytical skills from the model.
See Appendix \ref{Datasets} for more details.

\subsection{Baselines and Metrics}
We compared PG-RAG with a variety of advanced baseline methods: (1) GPT-3.5-turbo [25]; (2) native RAG methods including BM25 [26], dense passage retrieval (DPR) [23], hybrid retrieval [8], and hybrid retrieval with reranker [27]; (3) and RAG methods such as tree traversal [17], collapsed tree retrieval [17], Graph-RAG [28], and KGP [18]. More details in Appendix \ref{Baselines}.

In the experiments, two types of evaluation metrics are employed to assess the quality of generated text. Firstly, the overall semantic similarity metrics, including BLEU [29], ROUGE-L [30], and BertScore [31], evaluate how closely the generated content \(GM\) aligns with the reference content \(GT\) in terms of meaning and fluency. Secondly, the key information metric, RAGQuestEval [24], assesses how effectively the generated content captures and presents the key information from the reference content. Specifically, entities and noun phrases are extracted from the reference text to form a set of question-answer pairs \(\{(q,a)\}\) using a question generator \(Q_G(\cdot)\), where \(q\) represents a subquery and \(r\) the corresponding reference to verify whether the generated text contains and accurately conveys the content of the reference. We involved the \(F1_{QE}\) measure, which considers both precision and recall as follows:
\begin{equation}
    \operatorname{Prec}(GT, GM) = \frac{1}{\left|Q_G(GT)\right|} \sum_{(q, a) \in Q_G(GT)} F1\left(Q_A(GM, q), a\right)
    \label{eq:10},
\end{equation}
\begin{equation}
    \operatorname{Recall}(GT, GM) = \frac{1}{\left|Q_G(GT)\right|} \sum_{(q, a) \in Q_G(GT)} \mathbb{I}[Q_A(GM, q) \neq \operatorname{<Unanswerable>}]
    \label{eq:11}.
\end{equation}
where \(\mathbb{I}\) evaluates whether the response to a query \(q\) is "<Unanswerable>" or not.

\subsection{Implementation Details}
In all baseline methods, we follow the same RAG foundation settings as the CRUD-RAG benchmark, dividing the document repository into fixed-size chunks (128) and using a uniform embedding model (bge-base [23]) with a commonly used context window of 8. GPT-3.5 serves as the baseline model for the experiments, handling response generation and quality assessment. On this basis, we incorporate the RAG optimization modules proposed by each method. 
See Appendix \ref{Baselines} for more details.

\subsection{Results}

\textbf{Overall Comparison.} In Table~\ref{tab:retriever}, we compare the performance of our proposed PG-RAG method with other baseline methods, yielding the following results: 

 (1)  In terms of BERTScore and \(F1_{QE}\) metrics, the performance of the basic generative model lags behind that of methods using RAG. This is due to the model's knowledge base potentially being insufficient for such problems, leading to the generation of hallucinatory answers or choosing to evade, such as claiming, "I cannot provide news." The traditional RAG methods, integrating BM25 keyword search and DPR's deep semantic matching, can retrieve relevant information from documents more effectively. Among these, methods with a hybrid re-ranking mechanism perform the best within traditional RAG, as additional re-ranking algorithms can optimize the results of initial retrieval.
 
 (2) Tree-based baseline methods perform poorly across the three tasks, despite attempting to extract and summarize information from document blocks by constructing summary trees and trying to balance the performance of fine-grained (base blocks) and coarse-grained (summary) information retrieval through high-matching collapsed blocks or top-down adaptive search recall strategies. However, these baseline methods using the RAPTOR Cluster approach underperform in the large dataset used in this experiment, directly impacting the final results.

     \begin{wrapfigure}[16]{l}{0.45\textwidth} 
    \vspace{-4mm}
  \begin{center}  \includegraphics[width=0.45\textwidth]{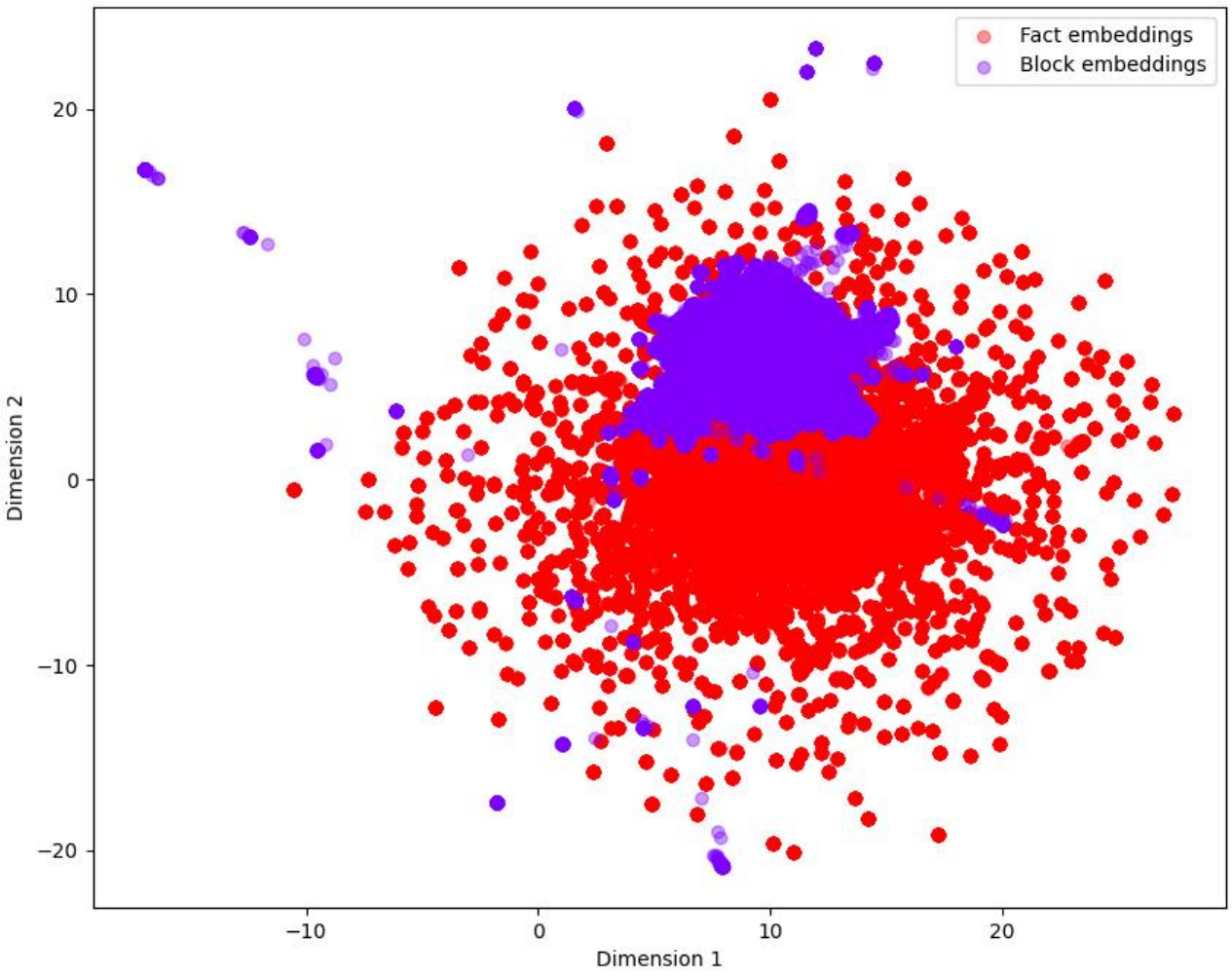}
  \end{center}
  \vspace{-2mm}
  \caption{Comparison of the block and fact embedding distributions in 2D space.}
  \label{fig:3}
\end{wrapfigure}
 (3) The Graph-RAG and KGP, two graph-based RAG methods that navigate the graph structure to recall evidence, outperform traditional RAG and tree-based RAG in multi-document tasks. Specifically, Graph-RAG uses beam search, while KGP relies on the reasoning capabilities of LLMs themselves, making adaptive evidence selection that meets the relationship reasoning demands of complex question answering. However, in single-document question answering, the performance improvement of these graph-based RAG models over traditional RAG is not significant due to the limited complexity and amount of information in single-document tasks.

 (4) Our PG-RAG method markedly surpasses all baselines in single-document question answering tasks, attributable to our transformation of the base text into a refined, well-structured, and highly coherent pseudo-graph structure. By navigating precisely through the routing nodes of the pseudo-graph, PG-RAG can focus on key information and effectively gather related context as a supplement. In tasks characterized by information redundancy and complexity across multiple documents, PG-RAG continues to demonstrate stable high performance.

\begin{wraptable}[12]{r}{0.40\textwidth}
\vspace{-2mm}
\centering
\caption{Comparison of three fusion algorithms based on the number of chunks to be fused, the number of clusters generated, and the average number of documents per cluster in topic fusion.}
\resizebox{0.40\textwidth}{!}{
\begin{tabular}{@{}lccc@{}}
\toprule
\textbf{Method} & \multicolumn{1}{c}{\textbf{Chunks}}                             & \multicolumn{1}{c}{\textbf{Clusters}}                         & \multicolumn{1}{c}{\begin{tabular}[c]{@{}c@{}}\textbf{Docs/}\\ \textbf{Cluster}\end{tabular}} \\ \midrule
RAPTOR & {153,577} & {46}    &  {277}                          \\
KNN    &  {153,577} &  {3,877} &  {2.69}                         \\
\textbf{Ours}   &  {10,436}  &  {3,500} &  {2.98}                         \\ \bottomrule
\end{tabular}
}
\label{tab:pg fusion}
\end{wraptable}

\textbf{Comparative Analysis of Knowledge Construction Methods.} Traditional RAG only performs simple chunking before retrieval. In contrast, tree-based and graph-based RAGs integrate these chunks through fusion algorithms, associating or aggregating them, but their performance highly depends on the quality and distribution of the initial segmentation. For instance, we segmented a dataset involving QA on 3,199 topics covering 10,436 articles, resulting in 153,577 chunk embeddings. As shown in Figure \ref{fig:3}, the embeddings obtained from our factual segmentation method are more dispersed in the embedding space compared to the original block embeddings, which benefits clustering. As shown in Table \ref{tab:pg fusion}, RAPTOR Cluster uses the GMM [32] algorithm to learn the distribution of embeddings awaiting clustering and adaptively determines the number of clustering layers. However, in large datasets, particularly where embedding distributions are concentrated, RAPTOR Cluster aggregates the 153,577 chunks into only 46 classes at the first layer, causing each summary block to aggregate information from nearly 227 documents, leading to overly abstract and noisy summary nodes. Additionally, RAPTOR Cluster assigns the blocks to be aggregated into a fixed number of learned clusters by setting thresholds. If the threshold is set too high, it becomes difficult for blocks to adapt to the clustering distribution, resulting in sparse connections between layers, and the final effect might be similar to ordinary chunking. If the threshold is too low, almost every block has a high probability of being integrated into the upper summary blocks, forming a full connection, which increases the noise in the relationships. Hence, the quality of the tree in RAPTOR Cluster is highly sensitive to the dataset and threshold settings. Graph-based methods mostly use KNN [33] for clustering. Here, we set the number of neighbors to 3, as the QA dataset was originally constructed by aggregating topics and selecting 1-3 documents from different topics to generate questions. We found that using KNN resulted in 3,877 clusters, close to the dataset’s own aggregation characteristics, but the fusion took too long, requiring nearly 5 hours with 20 threads set. Meanwhile, our method builds indices at the topic and fact levels, classifying each node with the top nodes meeting the threshold into one class, with a time complexity close to O(n). Searching for top nodes on the index is fast, in a single-thread setting, our fusion time was only 0.5 hours.

\begin{wraptable}[11]{r}{0.5\textwidth}
\vspace{-2mm}
\centering
\caption{Comparison of knowledge integration results using different RAG methods.}
\resizebox{0.5\textwidth}{!}{
\begin{tabular}{@{}cccc@{}}
\toprule
{ \textbf{Method}} & \begin{tabular}[c]{@{}c@{}}\textbf{Refinement}\\ \textbf{Rate}\end{tabular} & \begin{tabular}[c]{@{}c@{}}\textbf{Knowledge}\\ \textbf{Density}\end{tabular} & \begin{tabular}[c]{@{}c@{}}\textbf{Compression} \\ \textbf{Ratio} \end{tabular} \\
\midrule
Tree (\(\theta=0.3\)) & 31\% $\sim$ 32\% & $1.9 \cdot 10^{6}$ & 0.99 $\sim$ 1 \\
Tree (\(\theta=0.5\)) & 31\% $\sim$ 32\% & $1.55 \cdot 10^{6}$ & 0.99 $\sim$ 1 \\
Tree (\(\theta=0.8\)) & 31\% $\sim$ 32\% & $0.99 \cdot 10^{6}$ & 0.99 $\sim$ 1 \\KNN-Graph & 31\% & $4.88 \cdot 10^{6}$ & 1 \\
\textbf{PG (ours)} & 100\% & $4.11 \cdot 10^{6}$ & 2.64 \\

\bottomrule
\end{tabular}
}
\label{tab:pg analysis}
\end{wraptable}
\textbf{Comparative Analysis of Knowledge Construction Results.} As shown in Table \ref{tab:pg analysis}, it is clear that pseudo-graph generated post-fusion demonstrates excellent performance in terms of both the refinement level and density of knowledge. First, the graph comprises the highest number of nodes, each encapsulating distinct knowledge quantities and adjusting its granularity commensurate with the extent of the knowledge. Second, dense interconnections have been established among knowledge entities, facilitating the representation of various connection types reflective of the logical structure of knowledge. In contrast to the high-density KNN-Graph that employs a singular metric for all connections, our approach not only establishes essential links among knowledge entities but also minimizes connections between superficially similar yet fundamentally disparate knowledge pieces, thereby augmenting the quality. Crucially, the original knowledge has been effectively refined and compressed, achieving a compression ratio exceeding 2.5 times that of the original texts, thereby enhancing spatial efficiency.

    \begin{wrapfigure}[15]{l}{0.35\textwidth} 
    \vspace{-9mm}
  \begin{center}  \includegraphics[width=0.35\textwidth]{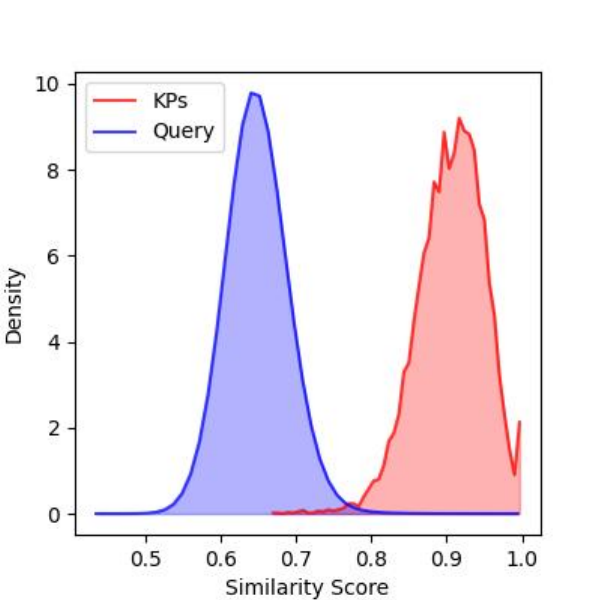}
  \end{center}
  \vspace{-3mm}
  \caption{Comparison of matching ability between raw query and KPs.}
  \label{fig:4}
\end{wrapfigure}

\textbf{Impact of Pseudo-Graph Retrieval.} PGR locates seed nodes in the initial recall sub-graph through KPs as evidence, which may be more powerful than using the original query directly for recall. As shown in Figure \ref{fig:4}, the distribution of similarity between KPs and the final answer (text composed of evidence set) is significantly higher than that between the original query and the final answer. This is because these key points can more precisely reflect the query intent, allowing for more direct location of relevant information when matching data in the knowledge base. 
Furthermore, assuming there are \( w \) candidate nodes, the complexity of performing a step-by-step DFS is \( O(w) \) because each reachable node needs to be visited sequentially. In PGR, \( w \) is stored within an \( m \times n = w \) matrix, requiring only 3 row traversals and 1 column traversal to complete the search. Hence, the complexity is \( O(m + 3n) \). Since the tree depth \( n \) is very small and the width \( m \) is large, in practical terms PGR outperforms DFS in terms of speed.

\section{Conclusion}
This paper proposes a pre-retrieval augmented generation method that introduces a "\textit{refinement}" step before the "\textit{indexing-retrieval-generation}" process, ensuring the accuracy of retrieved content from the outset. We leverage the self-learning capabilities of LLMs to transform documents into easily understandable and retrievable hierarchical indexes. This process naturally filters out noise and enhances information readability. By establishing connections between similar or complementary pieces of knowledge, we enable the retriever to function across multiple documents.
During the knowledge retrieval phase, we use "\textit{pseudo-answers}" to assist the retriever in locating relevant information and perform walking in the matrices, thereby achieving accurate and rapid knowledge localization. Finally, we assemble the retrieved fact paths into a structured context, providing rich background information for LLMs to generate knowledge-grounded responses.

\section*{References}
{
\small
[1] Kandpal, N., Deng, H., Roberts, A., Wallace, E., \& Raffel, C.\ (2023) Large language models struggle to learn long-tail knowledge. In \textit{International Conference on Machine Learning}, volume 202, pp.\ 15696--15707.

[2] Zhang, Y., Li, Y., Cui, L., Cai, D., Liu, L., Fu, T., Huang, X., Zhao, E., Zhang, Y., Chen, Y., Wang, L., Luu, A.T., Bi, W., Shi, F., \& Shi, S.\ (2023) Siren’s song in the AI ocean: A survey on hallucination in large language models. \textit{arXiv preprint arXiv:2309.01219}.

[3] Yang, K., Swope, A.M., Gu, A., Chalamala, R., Song, P., Yu, S., Godil, S., Prenger, R.J., \& Anandkumar, A.\ (2023) Leandojo: Theorem proving with retrieval-augmented language models. In \textit{Advances in Neural Information Processing Systems}.

[4] Mao, Y., Dong, X., Xu, W., Gao, Y., Wei, B., \& Zhang, Y.\ (2024) FIT-RAG: black-box RAG with factual information and token reduction. \textit{arXiv preprint arXiv:2403.14374}.

[5] Yan, S-Q., Gu, J-C., Zhu, Y., \& Ling, Z-H.\ (2024) Corrective retrieval augmented generation. \textit{arXiv preprint arXiv:2401.15884}.

[6] Liu, Y., Peng, X., Zhang, X., Liu, W., Yin, J., Cao, J., \& Du, T.\ (2024) RA-ISF: learning to answer and understand from retrieval augmentation via iterative self-feedback. \textit{arXiv preprint arXiv:2403.06840}.

[7] Langchain.\ (2023a) Parent document retriever. \url{https://python.langchain.com/v0.1/docs/modules/data_connection/retrievers/parent_document_retriever}.

[8] Langchain.\ (2023b) Ensemble retriever. \url{https://python.langchain.com/v0.1/docs/modules/data_connection/retrievers/ensemble}.

[9] Bratanič, T., \& Neo4j GenAI Research.\ (2024) Parent document retriever. \url{https://neo4j.com/developer-blog/knowledge-graph-rag-application}.

[10] Fatehkia, M., Lucas, J.K., \& Chawla, S.\ (2024) T-RAG: lessons from the LLM trenches. \textit{arXiv preprint arXiv:2402.07483}.

[11] Soman, K., Rose, P.W., Morris, J.H., Akbas, R.E., Smith, B., Peetoom, B., Villouta-Reyes, C., Cerono, G., Shi, Y., Rizk-Jackson, A., Israni, S., Nelson, C.A., Huang, S., \& Baranzini, S.E.\ (2023) Biomedical knowledge graph-enhanced prompt generation for large language models. \textit{arXiv preprint arXiv:2311.17330}.

[12] Kang, M., Kwak, J.M., Baek, J., \& Hwang, S.J.\ (2023) Knowledge graph-augmented language models for knowledge-grounded dialogue generation. \textit{arXiv preprint arXiv:2305.18846}.

[13] Zhang, H., Wang, Y., Chen, Q., Chang, R., Zhang, T., Miao, Z., Hou, Y., Ding, Y., Miao, X., Wang, H., Pang, B., Zhan, Y., Sun, H., Deng, W., Zhang, Q., Yang, F., Xie, X., Yang, M., \& Cui, B.\ (2023) Model-enhanced vector index. In \textit{Advances in Neural Information Processing Systems}.

[14] Langchain.\ (2023c) Semantic chunking. \url{https://python.langchain.com/v0.1/docs/modules/data_connection/indexing}.

[15] Berntson, A.\ (2023) Azure ai search: Outperforming vector search with hybrid retrieval and ranking capabilities. \url{https://techcommunity.microsoft.com/t5/ai-azure-ai-services-blog/azure-ai-search-outperforming-vector-search-with-hybrid/ba-p/3929167}.

[16] Chen, H., Pasunuru, R., Weston, J., \& Celikyilmaz, A.\ (2023) Walking down the memory maze: Beyond context limit through interactive reading. \textit{arXiv preprint arXiv:2310.05029}.

[17] Sarthi, P., Abdullah, S., Tuli, A., Khanna, S., Goldie, A., \& Manning, C.D.\ (2024) RAPTOR: recursive abstractive processing for tree-organized retrieval. \textit{arXiv preprint arXiv:2401.18059}.

[18] Wang, Y., Lipka, N., Rossi, R.A., Siu, A.F., Zhang, R., \& Derr, T.\ (2024) Knowledge graph prompting for multi-document question answering. In \textit{Thirty-Eighth AAAI Conference on Artificial Intelligence}, pp.\ 19206--19214.

[19] Chen, T., Wang, H., Chen, S., Yu, W., Ma, K., Zhao, X., Zhang, H., \& Yu, D.\ (2023) Dense X retrieval: What retrieval granularity should we use? \textit{arXiv preprint arXiv:2312.06648}.

[20] Wen, Y., Wang, Z., \& Sun, J.\ (2023) Mindmap: Knowledge graph prompting sparks graph of thoughts in large language models. \textit{arXiv preprint arXiv:2308.09729}.

[21] Gao, L., Ma, X., Lin, J., \& Callan, J.\ (2023) Precise zero-shot dense retrieval without relevance labels. In \textit{Proceedings of the 61st Annual Meeting of the Association for Computational Linguistics (Volume 1: Long Papers)}, pp.\ 1762--1777.

[22] Wu, Z., Pan, S., Chen, F., Long, G., Zhang, C., \& Yu, P.S.\ (2021) A comprehensive survey on graph neural networks. \textit{IEEE Transactions on Neural Networks and Learning Systems}, 32(1), pp.\ 4--24.

[23] Reimers, N., \& Gurevych, I.\ (2019) Sentence-bert: Sentence embeddings using siamese bert-networks. In \textit{Proceedings of the 2019 Conference on Empirical Methods in Natural Language Processing and the 9th International Joint Conference on Natural Language Processing}, pp.\ 3980--3990.

[24] Lyu, Y., Li, Z., Niu, S., Xiong, F., Tang, B., Wang, W., Wu, H., Liu, H., Xu, T., \& Chen, E.\ (2024) CRUD-RAG: A comprehensive chinese benchmark for retrieval-augmented generation of large language models. \textit{arXiv preprint arXiv:2401.17043}.

[25] Kalyan, K.S.\ (2023) A survey of GPT-3 family large language models including chatgpt and GPT-4. \textit{arXiv preprint arXiv:2310.12321}.

[26] Robertson, S.E., \& Zaragoza, H.\ (2009) The probabilistic relevance framework: BM25 and beyond. \textit{Found. Trends Inf. Retr.}, 3(4), pp.\ 333--389.

[27] FlagOpen.\ Flagembedding. \url{https://github.com/FlagOpen/FlagEmbedding}.

[28] NebulaGraph.\ (2024) Graph rag: Unleashing the power of knowledge graphs with llm. \url{https://medium.com/@nebulagraph/graph-rag-the-new-llm-stack-with-knowledge-graphs-e1e902c504ed}.

[29] Papineni, K., Roukos, S., Ward, T., \& Zhu, W-J.\ (2002) Bleu: a method for automatic evaluation of machine translation. In \textit{Proceedings of the 40th Annual Meeting of the Association for Computational Linguistics}, pp.\ 311--318.

[30] Lin, C-Y.\ (2004) ROUGE: A package

 for automatic evaluation of summaries. In \textit{Text Summarization Branches Out}, pp.\ 74--81. Barcelona, Spain: Association for Computational Linguistics.

[31] Zhang, T., Kishore, V., Wu, F., Weinberger, K.Q., \& Artzi, Y.\ (2020) Bertscore: Evaluating text generation with bert. In \textit{International Conference on Learning Representations}.

[32] Stauffer, C., \& Grimson, W.E.L.\ (1999) Adaptive background mixture models for real-time tracking. In \textit{Proceedings. 1999 IEEE Computer Society Conference on Computer Vision and Pattern Recognition (Cat. No PR00149)}, volume 2, pp.\ 246--252 Vol. 2.

[33] Cover, T.M., \& Hart, P.E.\ (1967) Nearest neighbor pattern classification. \textit{IEEE Trans. Inf. Theory}, 13(1), pp.\ 21--27.

[34] Guo, Z., Cheng, S., Wang, Y., Li, P., \& Liu, Y.\ (2023) Prompt-guided retrieval augmentation for non-knowledge-intensive tasks. In \textit{Findings of the Association for Computational Linguistics}, pp.\ 10896--10912.

[35] Asai, A., Wu, Z., Wang, Y., Sil, A., \& Hajishirzi, H.\ (2023) Self-rag: Learning to retrieve, generate, and critique through self-reflection. \textit{arXiv preprint arXiv:2310.11511}.

[36] Jeong, S., Baek, J., Cho, S., Hwang, S.J., \& Park, J.C.\ (2024) Adaptive-rag: Learning to adapt retrieval-augmented large language models through question complexity. \textit{arXiv preprint arXiv:2403.14403}.

[37] DataStax.\ (2024) How colbert helps developers overcome the limits of rag. \url{https://hackernoon.com/how-colbert-helps-developers-overcome-the-limits-of-rag}.

[38] McCormick, Z., Clark, J., \& Goldin, J.\ (2024) sprag. \url{https://github.com/SuperpoweredAI/spRAG}.

[39] Langchain.\ (2023d) Semantic chunking. \url{https://python.langchain.com/v0.1/docs/modules/data_connection/document_transformers/semantic-chunker}.

}

\appendix
\section{Limitations}
\label{limitations}
PG-RAG has three major limitations:
First, if the text is particularly long, LLMs may not be able to transform the extracted FCIs into a complete mind map due to their limited context window. To address this for long texts, we can pre-segment the text before PG construction to reduce the processing burden on LLMs. Although this method is feasible, the choice of pre-segmentation strategy can affect the organization of knowledge, and inappropriate segmentation may lead to the loss of contextual information.
Second, using LLMs for knowledge extraction is not economical for large-scale data scenarios, despite achieving satisfactory results. Therefore, further fine-tuning of lightweight models as alternatives is necessary. Fortunately, our two-stage extraction approach simplifies each step, allowing lightweight models to handle them. For example, converting text into fact-checking items is essentially a content rewriting process, while generating the mind map involves organizing the rewritten content. This drives us to continue model fine-tuning in future research.
Third, our walking algorithm performs only a simple exploration of the pseudo-graph, namely locating key points and walking to recall the context, failing to fully utilize the rich relationships contained in the pseudo-graph. For example, by leveraging the relationships between cross-document knowledge, we can compress redundant knowledge and merge complementary knowledge, effectively internalizing the information. Our initial work has not addressed this aspect, and future research will focus on designing walking algorithms to achieve knowledge pruning.
\section{Datasets}
\label{Datasets}
\begin{table}[b]
\centering
\caption{The summary of QA datasets.}
\label{tab:dataset}

\resizebox{\textwidth}{!}{
\begin{tabular}{@{}ccc@{}}
\toprule
Dataset Name                                                                               & Dataset Size           & Case                                                                                                                                                                                                                                                                                               \\ \midrule
\begin{tabular}[c]{@{}c@{}}Question Answering\\ (1-document)\end{tabular}                  & 3,199                  & \begin{tabular}[c]{@{}c@{}}Through observations with the Webb Space Telescope, researchers discovered the \\ presence of dust containing which element in galaxies less than a billion years old, \\ and this finding challenges which hypothesis about the formation of cosmic dust?\end{tabular} \\
\midrule
\multirow{4}{*}{\begin{tabular}[c]{@{}c@{}}Question Answering\\ (2-document)\end{tabular}} & \multirow{4}{*}{3,199} & \begin{tabular}[c]{@{}c@{}}In the first half of 2023, which was higher within Shanghai' jurisdiction: \\ the non-performing loan ratio in the banking industry or the original\\  insurance payout expenditure in the insurance industry?\end{tabular}                                             \\

                                                          \cmidrule{3-3}                                 &                        & \begin{tabular}[c]{@{}c@{}}How do the recent policy upgrades and service promotions by the State\\ Administration of Foreign Exchange specifically support the cross-border trade\\  and investment and financing of technology-based small and medium-sized enterprises?\end{tabular}             \\
                                                                                           \midrule
\multirow{4}{*}{\begin{tabular}[c]{@{}c@{}}Question Answering\\ (3-document)\end{tabular}} & \multirow{4}{*}{3,199} & \begin{tabular}[c]{@{}c@{}}In terms of ensuring food safety, what measures have the customs authorities, provincial\\ food safety committees, and local market supervision bureaus taken respectively?\end{tabular}                                                                                \\  \cmidrule{3-3}
                                                                                           &                        & \begin{tabular}[c]{@{}c@{}}In 2023, what adjustments did the domestic refined oil prices undergo, and how were\\ these adjustments influenced by changes in international oil prices and domestic policies?\end{tabular}                                                                           \\ \bottomrule 
\end{tabular}
}
\end{table}
In Table \ref{tab:dataset}, we provide several question-answering cases across different document types, totaling 3,199 entries per type.
\begin{itemize}
    \item \textbf{Question Answering (1-document):} includes questions that focus on fact-based information extracted from a single document. The questions require precise retrieval of specific details from the provided text.
    \item \textbf{Question Answering (2-document):} contains questions that necessitate the integration of information from two documents. The questions are designed to assess the ability to synthesize and compare information from multiple sources to arrive at a coherent answer.
    \item \textbf{Question Answering (3-document):} involves questions that require the extraction and synthesis of information from three documents. These questions test the model's capability to handle more complex queries, where relevant data must be gathered from multiple documents to form a complete and accurate response.
\end{itemize}

\section{Experiment Details}
\label{Baselines}
In this section, we summarize the key implementation details for each baseline as follows:

\begin{itemize}
    \item \textbf{GPT-3.5-turbo} [25] directly processes the questions from QA datasets and generates responses based solely on the input question without any external knowledge or context. This baseline helps assess the model's intrinsic ability to understand and respond to queries.
    
    \item \textbf{BM25} [26] is a sparse retrieval model that calculates the lexical overlapping score between the query and the text corpus. In our implementation, we divide the text into blocks. BM25 evaluates these blocks and ranks them based on the relevance scores. We then use the content from the top-8 blocks to answer the query. This approach tests the model's ability to retrieve relevant information based on lexical matching.
    
    \item \textbf{Dense Passage Retrieval} [23] uses a pre-trained language model \textit{bge-base} to embed text blocks. The query is also embedded using the same model, and the cosine similarity between the query vector and the text block vectors is computed. The top-8 blocks with the highest similarity scores are retrieved. This method evaluates the model's capability to find semantically relevant information using dense vector representations.
    
    \item \textbf{Hybrid Retrieval} [8] combines the strengths of BM25 and DPR by merging their retrieval results. We first retrieve blocks using both BM25 and DPR, creating a combined list of unique blocks. Each block is then scored using a Reciprocal Rank Fusion (RRF) approach, which calculates a weighted score based on the ranks from both retrieval methods. The blocks are sorted by these combined scores to determine the final ranking. The top-8 blocks with the highest similarity scores are retrieved. This hybrid approach leverages both lexical and semantic matching to improve retrieval accuracy.
    
    \item \textbf{Hybrid Retrieval with Reranker} [27] extends the hybrid retrieval method by adding a re-ranking step using the embedding model. After retrieving blocks using BM25 and DPR, we use the reranker model \textit{bge-rerank-base} to further refine the ranking of the top retrieved blocks. This reranker leverages advanced embedding techniques to enhance the accuracy of the final results. Then We choose the top-8 blocks as contexts. 
    
    \item \textbf{Tree Traversal} [17] starts at the root level of the summary tree and retrieves the top-8 nodes at each level based on cosine similarity to the query vector. At each level, it selects the top-8 nodes from the child nodes of the previous level's top-8 nodes. Finally, we choose the 8 nodes with the highest similarity as the context. This hierarchical method ensures that the most relevant nodes are progressively identified and refined as the traversal proceeds through the tree structure.
    
    \item \textbf{Collapsed Tree Retrieval} [17] flattens the tree structure into a single layer, ignoring the hierarchical structure. It calculates the cosine similarity of all nodes at all levels to the query vector and retrieves the top-8 nodes with the highest similarity as the context. This approach simplifies the retrieval process by treating the entire tree as a flat structure, allowing for more efficient identification of relevant nodes.

    \item \textbf{Graph-RAG} [28] uses KNN to cluster base blocks and constructs similarity links between nodes within the same cluster. During retrieval, we employ beam search with inner product as the scoring function to rank the blocks. Specifically, we limit the search depth to 3, as answering questions in our QA dataset typically requires up to 3 hops of reasoning. We set the number of blocks retrieved at each hop to 8, 5, and 3, respectively, generating a total of 120 retrieval paths. Each node's score along a path is the product of the scores from the first hop to the current hop. We rank all candidate nodes and select the top 8 as the final context.

    \item \textbf{KGP-LLaMA} [18] employs the LLaMA-7B model to generate the next supporting evidence and then selects neighboring nodes similar to this evidence as the next nodes to visit. Other settings are consistent with those of Graph-RAG.

    \item \textbf{PG-RAG (Ours)}. To ensure the quality and accuracy of the generated text, we set the similarity thresholds \( \theta_{bs} \) and \( \theta_{rl} \) at 0.85 and 0.75, respectively. This ensures that while compressing information, we retain key knowledge and maintain semantic coherence.
During the implementation of the walking algorithm, we set empirical values of 0.03 and 0.05 for the support threshold \( \theta_s \) and the fuzzy threshold \( \theta_f \). Setting lower thresholds means we are more stringent in selecting candidate nodes, thereby improving the accuracy of the final results. This method allows us to precisely filter out the most relevant candidate nodes.
Finally, we calculate the total contribution value of all candidate paths and select the top 8 fact paths with the highest total contribution values as our contextual information. These paths are typically within 128 tokens in length, as the content of the fact paths is closer to propositional expressions, making them more concise than basic blocks.

\end{itemize}

\begin{table}
\caption{Rleated works}
\resizebox{\textwidth}{!}{
\begin{tabular}{ccccccccc}
\toprule
\textbf{Method}                   &  \textbf{\begin{tabular}[c]{@{}c@{}}\textbf{Block}\\ \textbf{strategy}\end{tabular}}& \textbf{Block type} & \textbf{Data granularity}                                                                & \textbf{Data mode}                                                             & \textbf{\begin{tabular}[c]{@{}c@{}}Constraint\\ type\end{tabular}} & \textbf{\begin{tabular}[c]{@{}c@{}}Retrieval\\ dependency\end{tabular}} & \textbf{Walk} & \textbf{\begin{tabular}[c]{@{}c@{}}Retrieval\\ evidence\end{tabular}} \\ \toprule
{RAG} [8,34,37] & Standard          & Invariant           & Token/Sentence/Passage                                                           & Independent block                                                              & Hard                                                               & No                                                                      & No            & \begin{tabular}[c]{@{}c@{}}Independent\\ block\end{tabular}           \\
\midrule
             \begin{tabular}[c]{@{}c@{}}{RAG w.}\\ {overlap} [15]\end{tabular}               & Standard          & Incremental           & Token/Sentence/Passage                                                                 & Overlapping block                                                              & Hard                                                               & No                                                                      & No            & \begin{tabular}[c]{@{}c@{}}Independent\\ block\end{tabular}           \\
              \midrule
                  {PDR [7]}          & \begin{tabular}[c]{@{}c@{}}Standard or\\ semantic  \end{tabular}     & Incremental         & Granularity not uniform                                                                  & From small to large                                                            & Soft                                                               & Single                                                                  & No            & \begin{tabular}[c]{@{}c@{}}Chain tail\\ node\end{tabular}             \\
                 \midrule    \begin{tabular}[c]{@{}c@{}}{spRAG} [19, 38, 39]\end{tabular}       & Semantic          & *           & Granularity not uniform                                                                  & \begin{tabular}[c]{@{}c@{}}Independent block/\\ Overlapping block\end{tabular} & Soft                                                               & No or single                                                              & No            & \begin{tabular}[c]{@{}c@{}}Independent\\ block\end{tabular}           \\
                 \midrule
 \begin{tabular}[c]{@{}c@{}}{MEMWALKER [16];}\\ {RAPTOR [17]}\end{tabular}               & Semantic           & Incremental         & \begin{tabular}[c]{@{}c@{}}Block,\\ bolck summary\end{tabular}                   & Nodes and links                                                                & Soft                                                               & Single                                                                  & Yes           & \begin{tabular}[c]{@{}c@{}}Tree\\ nodes\end{tabular}              \\

 \midrule
{T-RAG [10]}                       & Semantic           & Compression         & Entities and triples                                                                             & Nodes and links                                                                & Hard                                                               & Single                                                                  & No            & Sub-tree                                                               \\
\midrule
 \begin{tabular}[c]{@{}c@{}}{KG-RAG [11];}\\ {SURGE} [12]\end{tabular}                     & Semantic        & Compression         & Entities and triples                                                                             & Nodes and links                                                                & Hard                                                               & Preset                                                                  & No            & Sub-graph                                                              \\
\midrule
{Mindmap [20]}                     & Semantic        & Compression         & Entities and triples                                                                             & Nodes and links                                                                & Hard                                                               & Preset                                                                  & Yes           & Sub-graph                                                              \\

\midrule
{KGP [18]}                         & Standard        & Invariant           & \begin{tabular}[c]{@{}c@{}}Block or\\ block structure\end{tabular}                & Nodes and links                                                                & Hard                                                               & Preset                                                                  & Yes           & Sub-graph                                                              \\
\midrule
{PG-RAG}                      & Semantic         & Compression         & \begin{tabular}[c]{@{}c@{}}Topics, routes, facts,\\ fact paths, concepts, etc.\end{tabular} & Nodes and links                                                                & Free                                                               & Various                                                          & Yes           & \begin{tabular}[c]{@{}c@{}}Pseudo\\ Sub-graph\end{tabular}             \\ \hline
\end{tabular}
}
\label{tab:rw}
\end{table}

\section{Related Work}
\label{Related Work}
\subsection{Index Construction for Efficient Information Retrieval}
In the field of information retrieval, index construction is a crucial step that directly impacts retrieval efficiency and accuracy. Traditional RAG (Retrieval-Augmented Generation) strategies enhance retrieval efficiency through different segmentation methods, such as ordinary segmentation and semantic segmentation. Ordinary segmentation strategies include independent blocks and overlapping blocks. Independent blocks divide data into fixed-size blocks that have no relationship with each other, while overlapping blocks allow overlap between blocks, increasing processing flexibility. These segmentation methods fall under hard constraint RAG approaches, where data patterns and granularity strictly adhere to predefined rules. In contrast, semantic segmentation adapts to the semantic content of the text with inconsistent granularity, falling under soft constraint RAG methods. It relaxes granularity constraints but remains limited to independent or overlapping blocks in most traditional RAG methods.

MEMWALKER and RAPTOR strategies extract and integrate paragraphs and paragraph summaries, relaxing granularity constraints and using the tree structure formed during integration for subsequent retrieval, falling under incremental segmentation with significant resource overhead. T-RAG employs a compressed segmentation strategy, extracting entities such as departments and organizations and the subordinate relationships forming an entity tree. It represents knowledge through nodes and showcases hierarchical and subordinate relationships between entities through links, providing a structured representation method for hierarchical information within organizations. When responding to user queries, T-RAG not only retrieves text blocks from the document database as context but also checks if the query mentions entities in the entity tree, thereby offering more accurate and enriched answers.

KG-RAG, SURGE, and Mindmap strategies achieve semantic segmentation by compressing data onto a structured knowledge graph. This hard constraint RAG paradigm ensures data constraints and entity relationship formation patterns, which are crucial for constructing accurate indexes and enhancing retrieval efficiency. However, the construction difficulty is extremely high, making it difficult to directly apply to new domain data. The KGP algorithm employs a general knowledge graph construction method, combining hard segmentation and similarity relationships to enhance the generalization ability of the knowledge graph. It divides data into paragraph or document structure blocks and calculates semantic similarity and keyword similarity between blocks to establish similarity links, relaxing data constraints, but the expressive capability of a single semantic similarity model is limited.

Our PG-RAG adopts an unconstrained semantic segmentation strategy, adaptively dividing data into themes, routes, facts, fact paths, concepts, events, and other granularities, allowing the construction of more dynamic and adaptable indexes to handle more complex retrieval tasks. PG-RAG adaptively segments according to knowledge granularity, and the relationships between nodes are not limited to subordinate relationships. They can also include parallel relationships, associative relationships, logical relationships, classification relationships, and expansion relationships, which are automatically determined based on the actual situation and needs of the data. This diversification of node relationships in the index construction method makes it more flexible and adaptable.
\subsection{Evidence Retrieval Mechanisms in Information Retrieval}
In the field of information retrieval, an effective evidence retrieval mechanism is crucial for enhancing retrieval quality and efficiency. Traditional Retrieval-Augmented Generation (RAG) strategies typically recall the top few independent blocks with the highest similarity as evidence, without leveraging relationships between blocks. RAG strategies based on incremental data models additionally incorporate simple hierarchical relationships between small and large blocks. During evidence retrieval, these models return the tail node of the chain as evidence, providing more comprehensive contextual information. Neither of these strategies involves complex walking mechanisms for retrieval.

MEMWALKER, on the other hand, relies on similarity relationships between summaries for contextual walking and retrieval. By using LLM-based agents, it makes top-down decisions on the summary tree and recalls the specific paragraphs pointed to by the summary chain that meet the requirements. This shows that MEMWALKER can navigate through the tree structure via specific paths to reach more detailed information points. RAPTOR employs both top-down and folded-tree search strategies for evidence retrieval.
T-RAG focuses on specific branches within the tree structure to extract relevant information. Specifically, T-RAG relies on subordinate relationships between entities during retrieval, directly returning the sub-tree formed by the path containing the required entities as evidence without performing additional navigation to recall more contextual information.

RAG methods based on (KGs, such as KG-RAG, SURGE, and Mindmap, generally use predefined relationships as retrieval dependencies. For example, KG-RAG presets connections between entities such as genes, proteins, drugs, compounds, and diseases. KG-RAG and SURGE perform sub-graph matching through entity linking, while Mindmap further expands the context using path-based and neighbor-based exploration strategies. Ultimately, these KG-based RAG methods use the retrieved sub-graph structure as evidence to assist the large model in response generation.
Our PG-RAG relies on adaptively generated rich relationships for the navigation and selection of evidence pseudo-graphs. This strategy is not limited to predefined relationship models but instead utilizes the inherent connections within the data itself for subsequent retrieval, demonstrating strong adaptability.

\section{Knowledge Fusion Algorithm}
\label{knowledge fusion algorithm}
\begin{algorithm}
\caption{Knowledge Fusion for Pseudo-Graph Generation}
\begin{algorithmic}[1]
\Require Mind maps $\mathcal{M}$, topic entities $V_t$, fact entities $V_f$.
\Ensure A pseudo-graph $\mathcal{PG}$.

\State Initialize visited nodes set $visited\_nodes \gets \{\}$
\For{each $node=$ $v_i\in V_{i=t, f}$}
    \If{$node.ID \not\in visited\_nodes$}
        \State $sim\_nodes \gets \text{findSimilarNodes}(node, V_i)$
        \State $super\_node \gets \text{createSuperNode}(sim\_nodes)$
        \State $\mathcal{PG}.\text{addEdge}(sim\_nodes, \text{link}, super\_node)$
        \State $visited\_nodes.\text{add}(sim\_nodes.ID)$
    \EndIf
\EndFor

\State \Return $\mathcal{PG}$

\end{algorithmic}
\label{al:pgg}
\end{algorithm}

    Algorithm \ref{al:pgg} outlines the knowledge fusion algorithm used for pseudo-graph generation. It accepts a collection of mind maps as input, comprising various topics, routes, and fact nodes, with topic nodes represented by topic vectors and fact nodes by path vectors. The algorithm begins by initializing two collections: one for storing the final cluster results and another for tracking visited and processed nodes. Subsequently, it commences a general processing sequence, traversing nodes of both topic and fact types. For each type of node, the algorithm verifies whether it has been visited. If not, it selects the appropriate vector set based on the node type, either the topic vector set \(T_v\) or the fact path vector set \(P_v\) and conducts a search for similar nodes. The function \text{findSimilarNodes}, which calculates similarity based on the node's vector and its comparison with other vectors in the vector set, identifies nodes that exceed a specific threshold. After identifying these similar nodes, the algorithm generates a new super-node that represents a group of closely related similar nodes. Subsequently, it establishes an edge labeled as a \textit{similarity link} between each identified similar node and the newly generated super-node. These links demonstrate that these nodes are similar in content and are now affiliated with a common super-node. The IDs of each identified similar node are added to the \(visited\_nodes\) collection, signifying that they have been processed. Meanwhile, the ID of the current node and its associated super-node are incorporated into the \text{Clusters} cluster collection. This process is executed identically for both topic and fact nodes. Finally, upon processing all nodes, the algorithm returns the constructed pseudo-graph PG. The constructed pseudo-graph PG not only aids in maintaining the organization of information but also enhances the intuitiveness and speed of information retrieval. Users can rapidly grasp the core content of specific clusters by exploring these super-nodes and further navigate to related detailed information through these nodes.

\section{Case Study}
\label{Case Study}
\begin{figure}
  \centering
  \includegraphics[width=\linewidth]{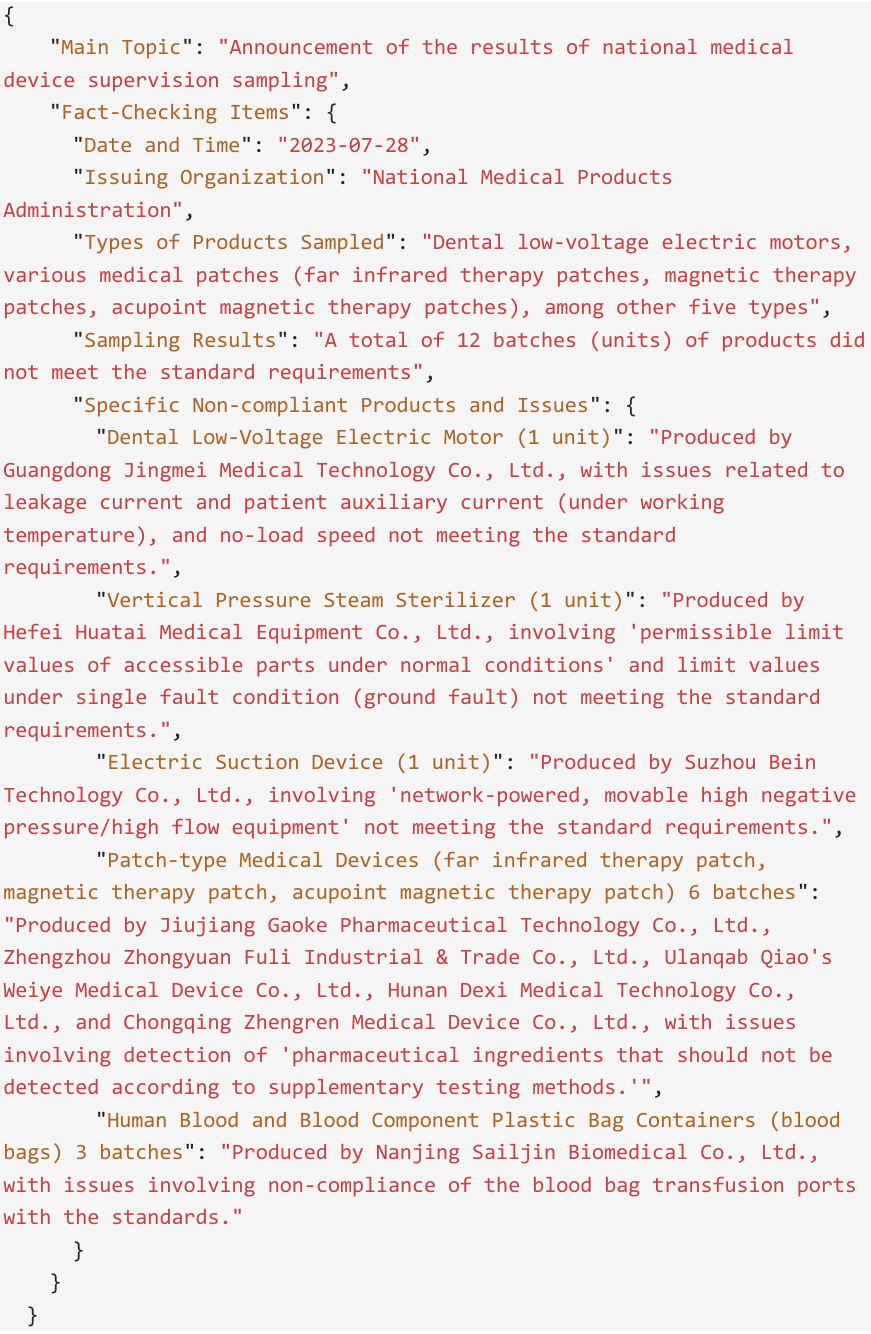}
  \caption{A specific case of main topic and fact-checking items.
}
\label{fig:fci}
\end{figure}

\begin{figure}
  \centering
  \includegraphics[width=\linewidth]{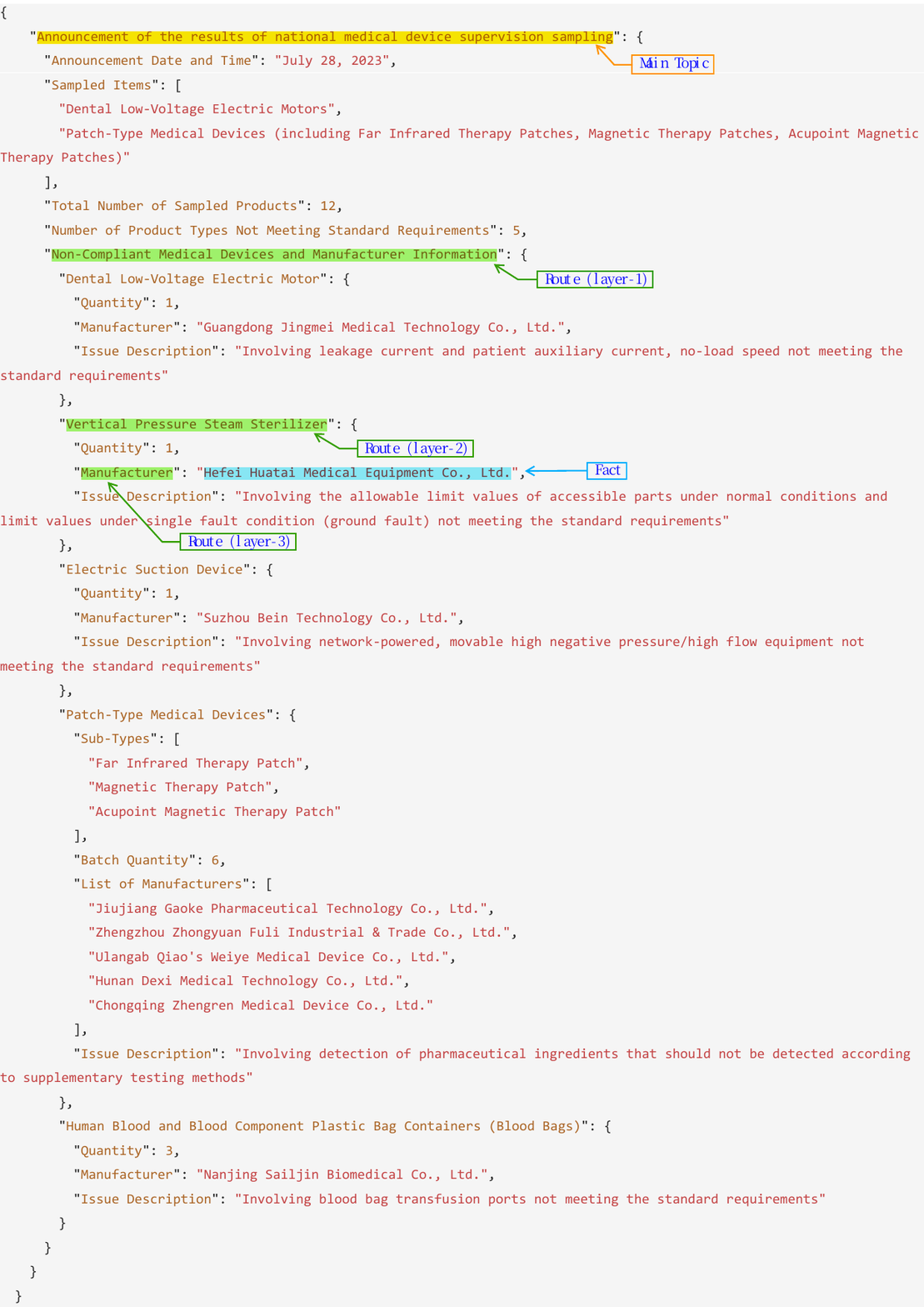}
  \caption{A specific case of the mind map generated through main topic and fact-checking items.
}
\label{fig:mindmap}
\end{figure}

In Figures \ref{fig:fci} and \ref{fig:mindmap}, we demonstrate the two-stage knowledge extraction process. Figure 1 shows that through pre-extraction of knowledge, we can preliminarily organize the information. Then, a mind map index is generated, which is concise, hierarchical, and effectively compresses the knowledge. Furthermore, we generate key answer points for queries. Specifically, we prompt the LLMs: "Please provide a concise and accurate answer key, listing the key or specific information needed to answer this question as comprehensively as possible." For instance, for a simple factual question like "What is the trend of dry eye occurrence?", the answer key would be "The trend of dry eye occurrence is increasing or decreasing." We retrieve similar information from a high-quality knowledge base, and if the fact indicates an increase, the answer trend will be increasing, and vice versa. For complex queries involving multiple answer points, such as "What environmental factors might exacerbate children's dry eye symptoms?", the model generates answer points from multiple perspectives, such as screen time and indoor air quality. We can concurrently locate relevant knowledge in the knowledge base, ensuring the breadth of the answer. Additionally, to enhance the depth of the answer, we perform PGR (Pseudo-Graph Retrieval) walks in the knowledge base to gather useful context. For example, for the question "What flood control measures has Beijing implemented?", we can locate the seed node, i.e., the fact path "Beijing -> flood control measures -> work stoppage", and then find the routing node "flood control measures" upwards to supplement more comprehensive knowledge. For more detailed questions like "What is the difference between work stoppage and production stoppage in flood control measures?", the fact path "Beijing -> flood control measures -> production stoppage" has a higher support probability than the fact path "Beijing -> flood control measures -> class suspension" and is used as a seed node to supplement knowledge, supporting the answer to the question.

\end{document}